\pdfoutput=1

\documentclass[11pt]{article}
\usepackage{acl}
\usepackage{times}
\usepackage{latexsym}
\usepackage[T1]{fontenc}
\usepackage[utf8]{inputenc}
\usepackage{microtype}
\usepackage{inconsolata}
\usepackage{graphicx}
\usepackage{booktabs}
\usepackage{multirow}
\usepackage{multicol}
\usepackage{subcaption}
\usepackage{url}

\usepackage{tcolorbox}
\usepackage{xcolor}

\usepackage{pifont}
\newcommand{\xmark}{\ding{55}}%

\usepackage{array} 
\usepackage{arydshln} 
\usepackage{subcaption}
\usepackage{danudefs}
\usepackage{acronym}

\usepackage{etoolbox}
\makeatletter
\newif\if@in@acrolist
\AtBeginEnvironment{acronym}{\@in@acrolisttrue}
\newrobustcmd{\LU}[2]{\if@in@acrolist#1\else#2\fi}

\newcommand{\ACF}[1]{{\@in@acrolisttrue\acf{#1}}}
\makeatother

\acrodef{MLM}[MLM]{Masked Language Model}
\acrodefplural{MLM}[MLMs]{\LU{M}{m}asked \LU{L}{l}anguage \LU{M}{m}odels}
\acrodef{LLM}[LLM]{Large Language Model}
\acrodefplural{LLM}[LLMs]{Large Language Models}
\acrodef{SoTA}[SoTA]{state-of-the-art}
\acrodef{ICL}{\LU{I}{i}n-cotext \LU{L}{l}earning}
\acrodef{SCD}{\LU{S}{s}emantic \LU{C}{c}hange \LU{D}{d}etection}
\acrodef{WiC}{Word-in-Context}
\acrodef{ITML}[ITML]{Information-Theoretic Metric Learning}
\acrodef{SDML}[SDML]{Semantic Distance Metric Learning}
\acrodef{CWE}{\LU{C}{c}ontextualised \LU{W}{w}ord \LU{E}{e}mbedding}
\acrodefplural{CWE}[CWEs]{\LU{C}{c}ontextualised \LU{W}{w}ord \LU{E}{e}mbeddings}
\acrodef{SCWE}[SCWE]{Sense-aware Contextualised Word Embedding}
\acrodefplural{SCWE}[SCWEs]{\LU{S}{s}ense-aware \LU{C}{c}ontextualised \LU{W}{w}ord \LU{E}{e}mbeddings}
\acrodef{PCA}[PCA]{Principal Component Analysis}
\acrodef{ICA}[ICA]{Independent Component Analysis}
\acrodef{TSP}[TSP]{Travelling Salesman Problem}
\acrodef{SWE}{\LU{S}{s}tatic \LU{W}{w}ord \LU{E}{e}mbedding}
\acrodefplural{SWE}[SWEs]{\LU{S}{s}tatic \LU{W}{w}ord \LU{E}{e}mbeddings}
\acrodef{WIT}[WIT]{Word Intruder Test}
\acrodef{SGNS}[SGNS]{Skip-Gram with Negative Sampling}
\usepackage[english]{babel}
\usepackage{hyperref}
\addto\captionsenglish{%
}
\addto\extrasenglish{%
}

\title{SCDTour: Embedding Axis Ordering and Merging for \\Interpretable Semantic Change Detection}

\author{Taichi Aida \\
  Tokyo Metropolitan Unviersity \\
  \texttt{taichia@tmu.ac.jp} \\\And
  Danushka Bollegala \\
  University of Liverpool \\
  \texttt{danushka@liverpool.ac.uk} \\}

\begin{document}
\maketitle
\begin{abstract}
In \ac{SCD}, it is a common problem to obtain embeddings that are both interpretable and high-performing.
However, improving interpretability often leads to a loss in the \ac{SCD} performance, and vice versa.
To address this problem, we propose \ac{SCD}Tour, a method that orders and merges interpretable axes to alleviate the performance degradation of \ac{SCD}.
\ac{SCD}Tour considers both (a) semantic similarity between axes in the embedding space, as well as (b) the degree to which each axis contributes to semantic change.
Experimental results show that \ac{SCD}Tour preserves performance in semantic change detection while maintaining high interpretability. 
Moreover, agglomerating the sorted axes produces a more refined set of word senses, which achieves comparable or improved performance against the original full-dimensional embeddings in the \ac{SCD} task.
These findings demonstrate that \ac{SCD}Tour effectively balances interpretability and \ac{SCD} performance, enabling meaningful interpretation of semantic shifts through a small number of refined axes.\footnote{Source code is available at \url{https://github.com/LivNLP/svp-tour}.}
\end{abstract}

\section{Introduction}
\label{sec:intro}
The meanings of words shift over time due to changes in culture, society, and contexts. 
\ac{SCD} is the task of detecting these changes automatically, which plays a vital role in aiding linguistic analysis~\cite{kutuzov-etal-2018-diachronic,schlechtweg-etal-2020-semeval}.
Additionally, it also contributes to the efficient additional training of \acp{MLM} by identifying words whose meanings have changed over time~\cite{Su-etal-2022-improving}.
A wide range of methods have been proposed using static ~\cite{kim-etal-2014-temporal,hamilton-etal-2016-diachronic,dubossarsky-etal-2019-time,aida-etal-2021-comprehensive} and contextualised word embeddings~\cite{kutuzov-giulianelli-2020-uio,rosin-etal-2022-time,rosin-radinsky-2022-temporal,aida-bollegala-2023-unsupervised,aida-bollegala-2023-swap} to enhance the performance of \ac{SCD}.

Recently, interpretability has emerged as a key focus in \ac{SCD}.
While earlier approaches prioritised improving accuracy using contextualised or static word embeddings, recent work has focused on transparency by generating definitions~\cite{giulianelli-etal-2023-interpretable,kutuzov-etal-2024-enriching-word}, building usage graphs~\cite{schlechtweg-etal-2021-dwug,ma-etal-2024-graph}, analysing embedding space structure~\cite{nagata-etal-2023-variance,aida-bollegala-2023-unsupervised}, or leveraging external knowledge~\cite{tang-etal-2023-word,periti-etal-2024-analyzing,baes-etal-2024-multidimensional}.
However, a key challenge remains: \textbf{improving interpretability often leads to reduced performance, and vice versa}~\cite{aida-bollegala-2025-investigating}.
This trade-off limits practical applications that demand both reliable predictions and interpretable explanations.\footnote{We define interpretability in \ac{SCD} as the ability to assign human-interpretable meanings to individual embedding axes. Unlike methods that generate textual definitions or use external knowledge bases, we focus on making each dimension of the representation space interpretable, enabling direct inspection of the semantic properties captured by the model.} 

\begin{table}[t]
    \centering
    \small
    \begin{tabular}{lcccc} \toprule
        Method & Dim. & Sorted & Perf. & Int.  \\ \midrule
        Raw & Full ($d$) & \xmark & \checkmark & \xmark \\ \hdashline
        PCA & Top-$k$ & eigenvalue & \checkmark & \xmark \\ 
        ICA & Top-$k$ & skewness & \xmark & \checkmark \\
        \textbf{SCDTour} & Merge-$k$ & TSP  & \checkmark & \checkmark \\ \bottomrule
    \end{tabular}
    \caption{Comparison of different embedding methods in terms of dimension (\textbf{Dim.}), axis-sorting strategy (\textbf{Sorted}), SCD performance (\textbf{Perf.}), and axis interpretability (\textbf{Int.}). Merge-$k$ represents the process of merging adjacent axes based on TSP sorting to obtain $k$-dimensional embeddings ($k < d$).}
    \label{tab:introduction}
    \vspace{-2em}
\end{table}

We address this issue by utilising interpretable embeddings whose axes are obtained via \ac{ICA}.
\ac{ICA} has been used to derive interpretable axes in word embeddings that encode meaning-specific information~\cite{yamagiwa-etal-2023-discovering}. 
We propose \textbf{SCDTour}, an interpretable axis-sorting method that extends prior work~\cite{sato-2022-wordtour,yamagiwa-etal-2024-axistour} by introducing \textit{change-specific weights} as a novel criterion, in addition to meaning-specific weights, to investigate \textbf{whether the \ac{ICA}-derived axes capture and explain semantic change of words}.
SCDTour enables us to sort and merge axes into interpretable embeddings while preserving \ac{SCD} performance.
Experimental results show that SCDTour can obtain low-dimensional, high-performing, and interpretable representations for \ac{SCD} against standard dimension reduction methods such as PCA and ICA (\autoref{tab:introduction}).

\begin{table}[t]
    \centering 
    \small
    \begin{tabular}{lrrr} \toprule
        Method  & Categorical & Similarity & Analogy \\ \midrule
        \textbf{Raw}, $d=300$ &  0.68 & 0.57 & 0.50 \\
        \multicolumn{4}{l}{\textbf{PCA}} \\
            \quad$k=5$ & 0.36 & 0.15 & \textbf{0.02} \\
            \quad$k=20$ & 0.49 & 0.23 & \textbf{0.09} \\
            \quad$k=100$ & 0.62 & 0.48 & 0.39 \\
        \multicolumn{4}{l}{\textbf{ICA$=$ICA(PCA)}} \\
            \quad$k=5$ & 0.30 & 0.06 & 0.00 \\
            \quad$k=20$ & 0.41 & 0.20 & 0.04 \\
            \quad$k=100$ & 0.60 & 0.46 & 0.35 \\ 
        \multicolumn{4}{l}{\textbf{PCA(ICA)}} \\
            \quad$k=5$ & 0.34 & 0.19 & 0.01 \\
            \quad$k=20$ &  0.42 & 0.39 & 0.04 \\
            \quad$k=100$ &  0.58 & \textbf{0.53} & 0.40 \\
        \multicolumn{4}{l}{\textbf{SCDTour} ($\lambda$ = 0.00)~\cite{yamagiwa-etal-2024-axistour}} \\
            \quad$k=5$ & \textbf{0.40} & \textbf{0.26} & 0.00 \\
            \quad$k=20$ & \textbf{0.52} & \textbf{0.42} & 0.07 \\
            \quad$k=100$ & \textbf{0.63} & 0.51 & \textbf{0.46} \\ \bottomrule
    \end{tabular}
    \caption{The performance of GloVe embeddings. We used the pretrained \texttt{GloVe 6B} model, referred to \citet{yamagiwa-etal-2024-axistour}. \textbf{ICA(PCA)} and \textbf{PCA(ICA)} indicate that PCA/ICA is conducted for the \textbf{Raw} embeddings to obtain full-dimensional axes ($d=300$), then ICA/PCA is performed to obtain $d$-dimensional embeddings.}
    \label{tab:benchmark} 
    \vspace{-1em}
\end{table}

\section{Method}
\label{sec:method}
We propose \textbf{\ac{SCD}Tour}, which introduces a \textbf{change-specific weight} to account for the contribution of each axis to \ac{SCD}.
Unlike WordTour~\cite{sato-2022-wordtour}, which sorts words based on pairwise similarity to obtain one-dimensional embeddings, and AxisTour~\cite{yamagiwa-etal-2024-axistour}, which aligns ICA axes and merges similar ones for better compression, our method incorporates change-specific signals to reorder and merge topic-like axes, thereby achieving both interpretability and \ac{SCD} performance.
\autoref{tab:introduction} shows that SCDTour is the only method that achieves both axis-level interpretability and high \ac{SCD} performance.

Previous work has explored how to sort axes in word embeddings to improve interpretability. 
WordTour~\cite{sato-2022-wordtour} reorders $n$ words in the $d$-dimensional original (Raw) \ac{SWE} $\mathbf{X} = [\vec{x}_1, \vec{x}_2, ..., \vec{x}_n]^{\mathsf{T}} \in \mathbb{R}^{n \times d}$.\footnote{WordTour focuses on constructing one-dimensional trajectories of words for interpretability. However, it is not directly applicable to \ac{SCD}, which requires comparing embeddings across multiple time periods.}
To obtain the optimal ordering $\sigma$, the task is formulated as a \ac{TSP} over words: 
\begin{equation}
    \min_{\sigma \in \mathcal{P}([n])} w(\sigma_1, \sigma_n) + \sum_{i=1}^{n-1} w(\sigma_i, \sigma_{i+1}),
    \label{eq:tour}
\end{equation}
where $\mathcal{P}([n])$ denotes the set of permutations of $n$ words, and $w(i, j)$ is a weight function that quantifies semantic distance between $i$-th and $j$-th words.
In WordTour, $w(i, j)$ is defined as the L1 distance between word embeddings $w(i, j) = \left|\left|\vec{x}_i - \vec{x}_j\right|\right|$.

Building on this idea, AxisTour~\cite{yamagiwa-etal-2024-axistour} proposed to sort \ac{ICA}-transformed axes $\mathbf{S} = \mathbf{A}_{\mathrm{ICA}}^{\mathsf{T}}\mathbf{X}^{\mathsf{T}} \in \mathbb{R}^{d \times n}$ by their semantic similarity between the $i$-th and $j$-th axes (\textbf{meaning-specific weight}) to obtain meaning-related ordering of axes:
\begin{equation}
    w_m(i, j) = \mathrm{cos}(\mathbf{v}_i, \mathbf{v}_j),
    \label{eq:weight_meaning}
\end{equation}
where $\mathbf{v}_i$ is the mean embeddings of the top $N$ words in the $i$-th axis. 
In addition, AxisTour introduced a dimension reduction technique ($\mathbb{R}^{n \times d} \rightarrow \mathbb{R}^{n \times k}$) by merging adjacent axes $I_r = \{a_r, .., b_r\}, r \in \{1, ..., k\}$ along the sorted order:
\begin{equation}
    f_r^{(\ell)} = 
    \begin{cases}
        \dfrac{\gamma_\ell^\alpha}{\sqrt{\sum_{i=a_r}^{b_r} \gamma_i^{2\alpha}}} & \text{for } \ell \in I_r, \\
        0 & \text{otherwise},
    \end{cases}
    \label{eq:dimension_reduction}
\end{equation}
where $\gamma_i$ denotes the skewness of the $i$-th axis, and $f_r^{(\ell)}$ indicates how much the $\ell$-th axis contributes to the $r$-th reduced dimension.

To extend this approach for semantic change detection (\ac{SCD}), we introduce an additional criterion: the \textbf{change-specific weight}, which evaluates how much each axis contributes to the performance on the \ac{SCD} task. 
Formally, we define it as:
\begin{equation}
    w_c(i, j) = \left|\left|Imp(j) - Imp(i)\right|\right|,
    \label{eq:weight_change}
\end{equation}
where $Imp(i)$ is the \textbf{importance of the $i$-th axis} compared to all other dimensions $\{D\}$, defined as below:
\begin{equation}
    Imp(i) = E(\mathbf{S}_{\{D\}}) - E(\mathbf{S}_{\{D\}\setminus\{i\}}).
    \label{eq:weight_change_importance}
\end{equation}
It quantifies the drop in performance $E$ when the $i$-th axis is removed.
Our proposed method, \textbf{\ac{SCD}Tour}, combines both criteria to produce a more informative and \ac{SCD}-relevant axis ordering:
\begin{equation}
    w(i, j) = \lambda w_c(i, j) + (1 - \lambda) w_m(i, j),
    \label{eq:weight_scdtour}
\end{equation}
where $\lambda \in [0, 1]$ is a hyperparameter to control the change-specific weight.
This generalises previous methods, which rely solely on semantic similarity.
While interpretability is not guaranteed, our preliminary experiment shows that this merging strategy achieves comparable or better performance to standard dimension reduction methods such as PCA or ICA on word embedding benchmarks (\autoref{tab:benchmark}).

\section{Experiments}
\label{sec:exp}

\subsection{Settings}
\label{subsec:exp_setting}
To evaluate the effectiveness of SCDTour, we focus on two aspects: (a) the interpretability of axes, and (b) its performance in \ac{SCD}.
Additionally, we investigate how the weighting parameter $\lambda$, balancing change-specific and meaning-specific weights in \autoref{eq:weight_scdtour}, influences both aspects.
Following previous work, we employ the LKH solver~\cite{HELSGAUN2000106} to solve the TSP formulated in \autoref{eq:tour}.

\paragraph{Interpretability:} To assess interpretability, we use the \ac{WIT}~\cite{musil-marecek-2024-exploring}. This task measures the axis coherence by introducing a semantically unrelated \textit{intruder} word into a set of related words\footnote{In this paper, we assume that the top-10 words in each axis serve as its representative words~\cite{yamagiwa-etal-2023-discovering,yamagiwa-etal-2024-axistour,musil-marecek-2024-exploring}.} and checks whether an evaluator can correctly identify the intruder word. Following \citet{musil-marecek-2024-exploring}, we use \acp{LLM} to simulate human-level evaluation. Specifically, we adopt three publicly available instruction-tuned models: Llama-3.1,\footnote{\url{https://huggingface.co/meta-llama/Llama-3.1-8B-Instruct}} Gemma-3,\footnote{\url{https://huggingface.co/google/gemma-3-4b-it}} and Qwen3.\footnote{\url{https://huggingface.co/Qwen/Qwen3-8B}}
Llama-3.1 has previously demonstrated effectiveness in a recent \ac{SCD} task~\cite{periti-etal-2024-automatically}. We also include Gemma-3 and Qwen3 to examine the robustness of our interpretability results across different LLM architectures and training strategies. We prompt the model in a zero-shot setting and postprocess outputs to extract a single-word prediction. To account for randomness in generation, we report the average accuracy over five runs.
\paragraph{Performance:} We use the standard benchmark, SemEval-2020 Task 1~\cite{schlechtweg-etal-2020-semeval}, which provides two time-separated corpora and a list of target words.\footnote{Dataset statistics are shown in \autoref{app_subsec:details_data}.} Following prior works~\cite{cassotti-etal-2023-xl,periti-etal-2024-analyzing,aida-bollegala-2024-semantic}, we mainly conduct the ranking task and measure the Spearman's correlation between semantic change scores and human ratings.
In addition, we also evaluate the binary classification setting, where the goal is to decide whether a target word has changed in meaning across time periods.
Similar to the WIT evaluation, we adopt three \acp{LLM}.
However, the binary setting is evaluated under few-shot prompting to mitigate overly strict decisions.\footnote{Prompt templates for both \ac{WIT} and binary \ac{SCD} are shown in \autoref{app_subsec:prompts}.}

In our experiments, we use \acp{SWE} instead of \acp{CWE}, which aligns with previous studies~\cite{sato-2022-wordtour,yamagiwa-etal-2024-axistour,musil-marecek-2024-exploring}, because \acp{SWE} provide more explicit access to axis-level information.
While \acp{CWE} encode sense-aware information for each token occurrence and achieve higher performance on \ac{SCD}~\cite{cassotti-etal-2023-xl}, it makes axis-level interpretation difficult due to the large number of contextualised instances.
In contrast, \acp{SWE} assign a single vector per word, allowing us to directly inspect which words dominate each axis.
We use \ac{SGNS} embeddings trained on time-separated corpora and apply Orthogonal Procrustes~\cite{hamilton-etal-2016-diachronic}.\footnote{We tune hyperparameters as described in \autoref{app_subsec:details_hyperparameters}.}
To compute the semantic change scores between time-separated embeddings, we use cosine similarity.

\begin{figure*}[t]
    \centering
    \begin{subfigure}{0.32\linewidth}
        \centering
        \includegraphics[width=\linewidth]{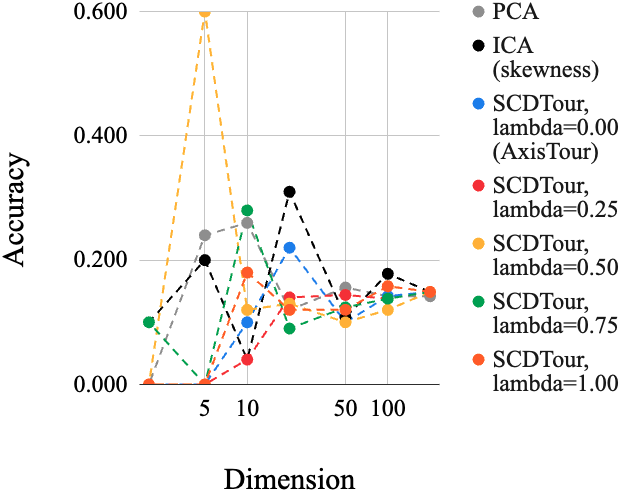}
        \caption{Llama-3.1}
        \label{subfig:wit_llama}
    \end{subfigure}
    \hfill
    \begin{subfigure}{0.32\linewidth}
        \centering
        \includegraphics[width=\linewidth]{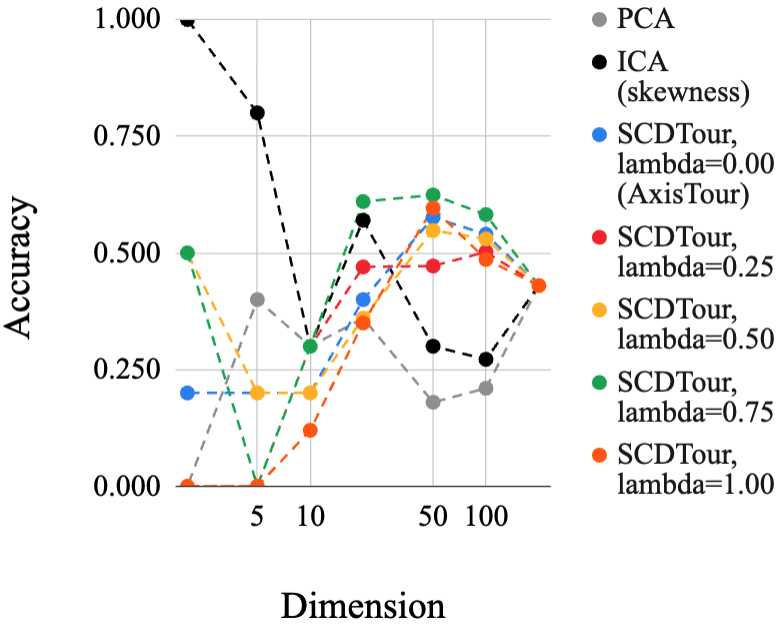}
        \caption{Gemma-3}
        \label{subfig:wit_gemma}
    \end{subfigure}
    \hfill
    \begin{subfigure}{0.32\linewidth}
        \centering
        \includegraphics[width=\linewidth]{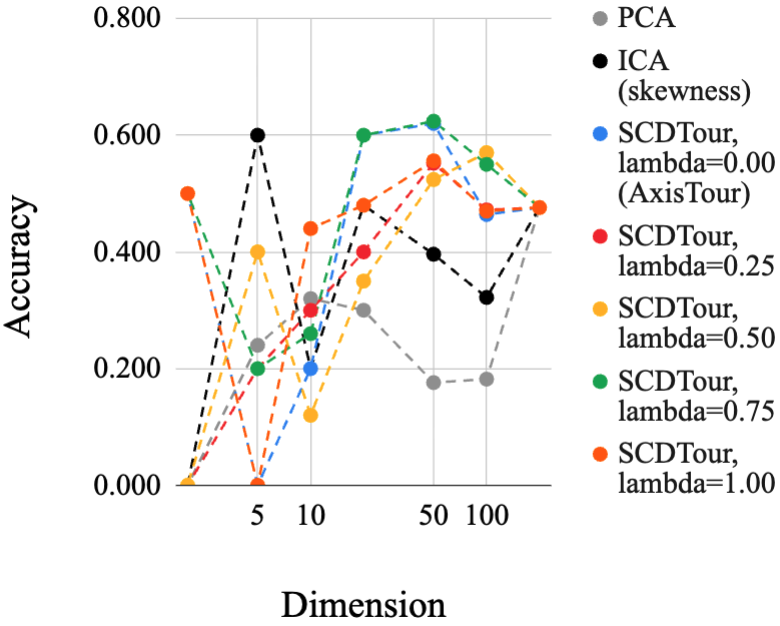}
        \caption{Qwen3}
        \label{subfig:wit_qwen}
    \end{subfigure}
    \caption{Accuracy on the word intruder test.}
    \label{fig:wit}
\end{figure*}

\begin{figure*}[t]
    \centering
    \begin{subfigure}{0.32\linewidth}
        \centering
        \includegraphics[width=\linewidth]{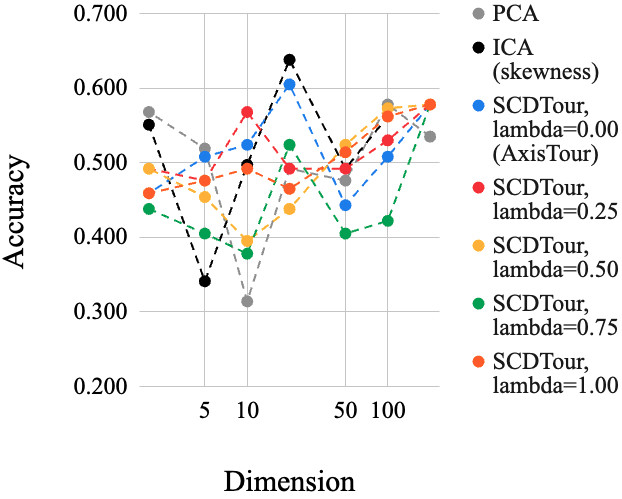}
        \caption{Llama-3.1}
        \label{subfig:scd_binary_llama}
    \end{subfigure}
    \hfill
    \begin{subfigure}{0.32\linewidth}
        \centering
        \includegraphics[width=\linewidth]{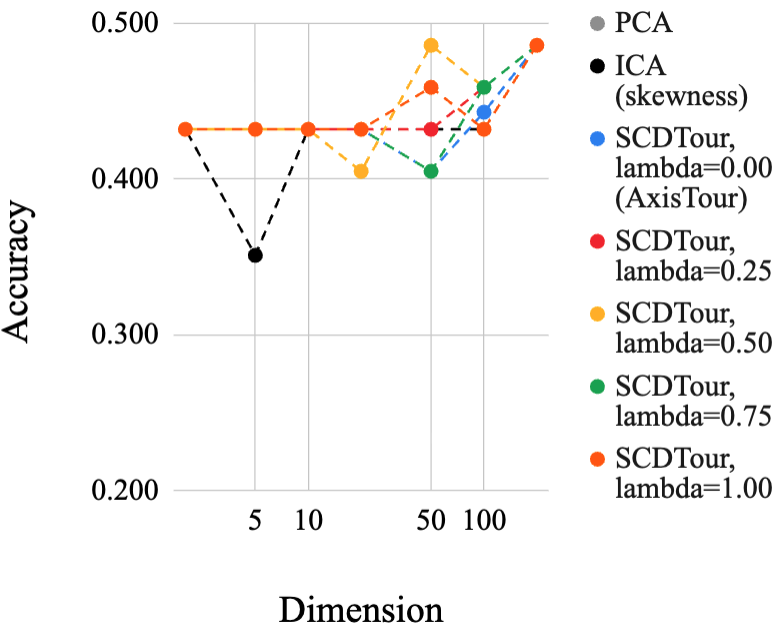}
        \caption{Gemma-3}
        \label{subfig:scd_binary_gemma}
    \end{subfigure}
    \hfill
    \begin{subfigure}{0.32\linewidth}
        \centering
        \includegraphics[width=\linewidth]{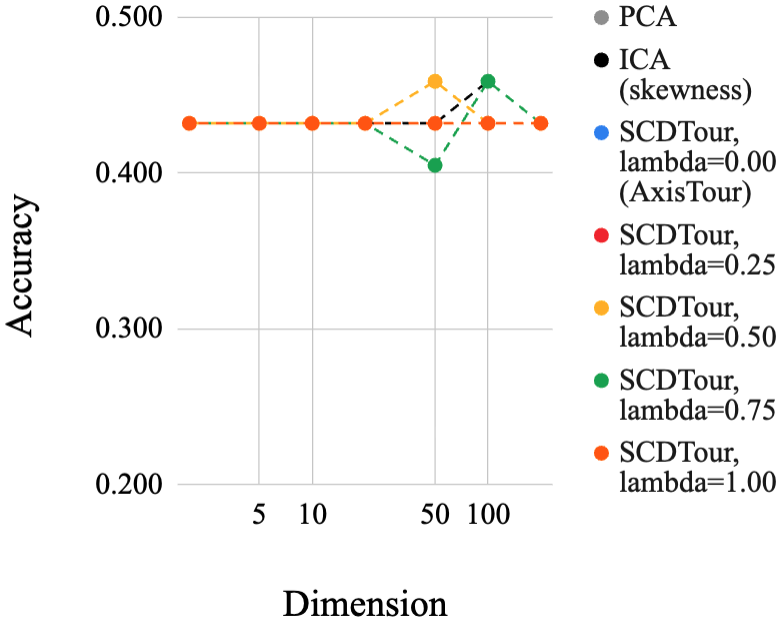}
        \caption{Qwen3}
        \label{subfig:scd_binary_qwen}
    \end{subfigure}
    \caption{Accuracy on the semantic change detection.}
    \label{fig:scd_binary}
\end{figure*}

\begin{figure}[t]
    \centering
    \includegraphics[width=\linewidth]{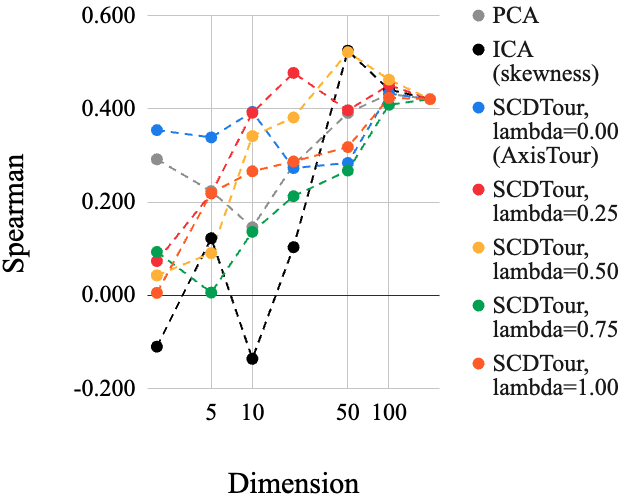}
    \caption{Spearman's rank correlation for the semantic change detection task.}
    \label{fig:scd_ranking}
    \vspace{-1em}
\end{figure}

\subsection{Results}
\label{subsec:exp_results}

\paragraph{RQ: Can \ac{SCD}Tour maintain interpretability with sorted/gathered \emph{sense} axes?}
\autoref{fig:wit} shows the accuracy on the \ac{WIT} using three \acp{LLM}.
Across all three models, SCDTour w/ the change-specific weight ($\lambda > 0$) performs comparably against ICA and SCDTour ($\lambda = 0$) for interpretability, confirming that our method maintains interpretability regardless of the underlying \ac{LLM}.
As shown by \citet{yamagiwa-etal-2023-discovering}, PCA fails to provide interpretable axes compared to ICA.

\begin{table*}[t]
    \centering
    \resizebox{\textwidth}{!}{
    \begin{tabular}{lrlrl} \toprule
        \multirow{2}{*}{Method} & \multicolumn{2}{c}{word: \textit{graft}} & \multicolumn{2}{c}{word: \textit{chairman}}  \\ 
        & axis & words & axis & words \\ \midrule
        \multicolumn{5}{c}{$d = 200$ (full)} \\ \midrule
            \multirow{2}{*}{Raw} & \textcolor{black!50}{70} & \textcolor{black!50}{\textbf{flowering}, \textbf{pear}, \textbf{pendulous}, \textbf{deciduous}, \textbf{sycamore}} & \textcolor{black!50}{42} & \textcolor{black!50}{\textbf{vote}, \textbf{election}, \textbf{senate}, \textbf{senator}, \textbf{delegate}} \\ 
             & 157 & \textbf{respiratory}, \textbf{intestinal}, \textbf{pulmonary}, \textbf{uterine}, \textbf{inflammation} & 42 & \textbf{caucus}, \textbf{senatorial}, \textbf{republican}, \textbf{democrat}, \textbf{nominee} \\ \midrule
            \multicolumn{5}{c}{$k = 100$} \\ \midrule
            \multirow{2}{*}{PCA} & \textcolor{black!50}{65} & \textcolor{black!50}{conceivable, precaution, agatha, believe, alteration} & \textcolor{black!50}{94} & \textcolor{black!50}{globe, self-satisfied, isaiah, dainty, area} \\
             & 66 & prop, ragusa, malignity, thieving, vanegas & 53 & \textbf{poky}, avenue, t1, tint, omnipotence \\
            \multirow{2}{*}{ICA} & \textcolor{black!50}{70} & \textcolor{black!50}{\textbf{pear}, \textbf{sycamore}, \textbf{shrubbery}, \textbf{tree}, \textbf{fern}} & \textcolor{black!50}{99} & \textcolor{black!50}{+, --, ditto, sauk, hydrochloric} \\
             & 82 & realist, condensed, t', misapply, commodity & 94 & rid, \textbf{accuse}, dispose, whiff, incapable \\
            \multicolumn{5}{l}{\textbf{SCDTour}} \\
            \multirow{2}{*}{\:$\lambda = 0.00$} & \textcolor{black!50}{73} & \textcolor{black!50}{\textbf{tree}, \textbf{sycamore}, \textbf{pendulous}, \textbf{pear}, \textbf{deciduous}} & \textcolor{black!50}{17} & \textcolor{black!50}{vote, election, presidential, elect, candidate} \\
             & 12 & \textbf{inflammation}, \textbf{intestinal}, \textbf{pulmonary}, \textbf{respiratory}, \textbf{disease} & 67 & leo, \textbf{authority}, \textbf{department}, quarter, 20th \\ 
            \multirow{2}{*}{\:$\lambda = 0.25$} & \textcolor{black!50}{41} & \textcolor{black!50}{\textbf{pear}, \textbf{elm}, hampshire, \textbf{shrub}, \textbf{flowering}} & \textcolor{black!50}{15} & \textcolor{black!50}{\textbf{vote}, \textbf{election}, \textbf{senator}, \textbf{senate}, \textbf{judiciary}} \\
             & 92 & \textbf{pulmonary}, \textbf{respiratory}, \textbf{infection}, \textbf{colon}, \textbf{liver} & 15 & \textbf{republican}, \textbf{caucus}, \textbf{congressional}, \textbf{senatorial}, \textbf{gubernatorial} \\
            \multirow{2}{*}{\:$\lambda = 1.00$} & \textcolor{black!50}{5} & \textcolor{black!50}{\textbf{flowering}, \textbf{sycamore}, \textbf{shrub}, \textbf{pear}, \textbf{herbaceous}} & \textcolor{black!50}{53} & \textcolor{black!50}{\textbf{vote}, \textbf{senator}, \textbf{election}, \textbf{representative}, \textbf{ballot}} \\
            & 80 & givin, intestinal, seein, \textbf{pulmonary}, \textbf{respiratory} & 12 & incapable, dozens, \textbf{accuse}, fond, kind \\ \midrule
        \multicolumn{5}{c}{$k = 20$} \\ \midrule
            \multirow{2}{*}{PCA} & \textcolor{black!50}{14} & \textcolor{black!50}{hutchinson, montague, rain, mosquito, kirk} & \textcolor{black!50}{10} & \textcolor{black!50}{disbelief, god, almighty, forgiveness, whosoever} \\
             & 14 & louisiana, piping, lilac, predominate, starch & 14 & louisiana, piping, lilac, predominate, starch \\
            \multirow{2}{*}{ICA} & \textcolor{black!50}{11} & \textcolor{black!50}{exempt, extricate, exemption, deviate, detract} & \textcolor{black!50}{14} & \textcolor{black!50}{address, customary, ducat, color, alice} \\
             & 13 & beyond, shirt, \textbf{medication}, determine, revolution & 17 & cease, lately, connection, lydia, already \\
            \multicolumn{5}{l}{\textbf{SCDTour}} \\
            \multirow{2}{*}{\:$\lambda = 0.00$} & \textcolor{black!50}{13} & \textcolor{black!50}{amends, precedence, necessary, precaution, observation} & \textcolor{black!50}{3} & \textcolor{black!50}{\textbf{indictment}, \textbf{politician}, \textbf{lunatic}, \textbf{defendant}, \textbf{adjudge}} \\
             & 2 & \textbf{transplant}, \textbf{inflammation}, \textbf{infection}, \textbf{disorder}, \textbf{respiratory} & 2 & transplant, inflammation, infection, disorder, respiratory \\ 
            \multirow{2}{*}{\:$\lambda = 0.25$} & \textcolor{black!50}{8} & \textcolor{black!50}{\textbf{pear}, \textbf{vine}, \textbf{tree}, \textbf{elm}, \textbf{pendulous}} & \textcolor{black!50}{3} & \textcolor{black!50}{\textbf{magistracy}, \textbf{elect}, curtis, \textbf{amendment}, \textbf{legislature}} \\
             & 18 & \textbf{respiratory}, chronic, \textbf{infection}, \textbf{pulmonary}, \textbf{renal} & 3 & \textbf{assembly}, \textbf{congressional}, driver, \textbf{nominee}, \textbf{caucus} \\
            \multirow{2}{*}{\:$\lambda = 1.00$} & \textcolor{black!50}{1} & \textcolor{black!50}{hemlock, stunted, moneywort, \textbf{crop}, \textbf{apple-tree}} & \textcolor{black!50}{10} & \textcolor{black!50}{artillery, picket, apartment, palace, portico} \\
            & 0 & 2,200, audible, sidle, syllable, difference & 2 & fond, \textbf{accuse}, let, plenty, faster \\ \midrule
        \multicolumn{5}{c}{$k = 5$} \\ \midrule
            \multirow{2}{*}{PCA} & \textcolor{black!50}{4} & \textcolor{black!50}{pie, mince, first-rate, stuff, tight} & \textcolor{black!50}{3} & \textcolor{black!50}{glasgow, 1835, 43, sloop, 1830} \\
             & 4 & tasty, \textbf{pill}, prescription, \textbf{medication}, dessert & 3 & kentucky, oakland, 153, md, fl \\
            \multirow{2}{*}{ICA} & \textcolor{black!50}{0} & \textcolor{black!50}{hebraic, ludicrousness, orvieto, tannin, wattie} & \textcolor{black!50}{3} & \textcolor{black!50}{qui, je, vous, comme, zo} \\
             & 0 & gainsay, hoyden, condi, monarchial, bb & 3 & sus, que, por, la, como \\
            \multicolumn{5}{l}{\textbf{SCDTour}} \\
            \multirow{2}{*}{\:$\lambda = 0.00$} & \textcolor{black!50}{3} & \textcolor{black!50}{militia, thornton, assistance, impression, opportunity} & \textcolor{black!50}{0} & \textcolor{black!50}{1794, roldan, misma, -, ce} \\
             & 3 & enormous, specimen, revelation, outstreched, handkerchief & 0 & ciudad, mrs, cell, marietta, \textbf{breed} \\ 
            \multirow{2}{*}{\:$\lambda = 0.25$} & \textcolor{black!50}{2} & \textcolor{black!50}{laboratory, deduce, imaginary, varmint, jury} & \textcolor{black!50}{2} & \textcolor{black!50}{laboratory, deduce, imaginary, varmint, jury} \\
             & 4 & \textbf{illness}, \textbf{dental}, m, nightgown, hem & 2 & clearly, greased, cultivated, legitimate, hamlet \\
            \multirow{2}{*}{\:$\lambda = 1.00$} & \textcolor{black!50}{3} & \textcolor{black!50}{cambridge, xx\_v, aime, mocha, tut} & \textcolor{black!50}{2} & \textcolor{black!50}{billows, manufactory, prevail, taste, jerry} \\
            & 0 & unison, credible, anticipated, \$800, dramatically & 0 & unison, credible, anticipated, \$800, dramatically \\ \bottomrule
    \end{tabular}
    }
    \caption{Representative words from the most activated axis of the target word embedding (\textit{graft} and \textit{chairman}) at two time periods $t_1$ (shown in \textcolor{black!50}{gray}) and $t_2$ (in black), across different methods. For each method and target word, we identify the axis with the highest value in the embedding, and list the top-$5$ words associated with that axis. Words that reflect the meaning of the target word are highlighted in \textbf{bold}.}
    \label{tab:analysis_graft_chairman}
    \vspace{-1em}
\end{table*}

\paragraph{RQ: Can \ac{SCD}Tour solve the \ac{SCD} task with the sorted/gathered axes?}
On the \ac{SCD} ranking task, \autoref{fig:scd_ranking} shows that SCDTour ($\lambda = 0.25, 0.50$) outperforms baselines (PCA, ICA, and SCDTour ($\lambda = 0.00$)) in the low- and mid-dimensional settings ($k = 20, 50, 100$).
While SCDTour ($\lambda = 0.00$) performs best in extremely low-dimensional settings ($k = 2, 5$), its interpretability is limited, as we discuss later.
\autoref{fig:scd_binary} shows results across three \acp{LLM}.
With Llama-3.1, SCDTour achieves the least degradation when reducing dimensionality, producing balanced judgments in few-shot settings.
In contrast, Gemma-3 and Qwen3 frequently defaulted to NO outputs across most configurations, leading to reduced discrimination.
This discrepancy may reflect differences in instruction-following capabilities.\footnote{LLaMA-3.1 has been fine-tuned using human feedback data, whereas the extent to which Gemma-3 and Qwen3 rely on synthetic/human-generated instruction data remains unclear.}
Such differences in training strategy appear to strongly influence semantic judgment tasks that require alignment with human intuition~\cite{sorensen-etal-2022-information}.

To further investigate the interpretability of the learned axes, we analyse two representative target words: \textit{graft}, which underwent a semantic change (from \textit{horticultural grafting} to \textit{medical transplant}), and \textit{chairman}, which maintained a stable meaning over time.
Results are shown in \autoref{tab:analysis_graft_chairman}.
We see that at $d = 200$ (full dimension) and $k = 100$, most methods can retrieve interpretable axes whose top-ranked words correspond to the meanings in each time period.
In contrast, PCA constantly fails to extract such axes even at $k = 100$, highlighting its limited utility for interpretability.
For $k = 20$, \ac{SCD}Tour ($\lambda = 0.25$) successfully identifies axes that capture each relevant meaning of the target words. 
However, SCDTour ($\lambda = 0$) fails to capture the axis representing the older meaning of \textit{graft}.
When the number of dimensions is reduced to 5, no method reliably produces axes with coherent representative words.
In such cases, the top words on each axis lack semantic consistency and fail to reflect interpretable meanings. 
We hypothesise that this is because the number of dimensions becomes insufficient to encode the full range of word meanings, causing multiple unrelated axes to be merged, according to \autoref{eq:dimension_reduction}.
A similar trend is observed for the qualitative analysis in \autoref{app_sec:scd}.

Overall, these findings demonstrate that the use of the change-specific weight enables SCDTour to efficiently maintain interpretability and performance, even with a reduced number of dimensions.

\section{Conclusion}
\label{sec:conclusion}
We presented SCDTour, a method that orders and merges interpretable axes using meaning- and change-specific weights. 
It maintains interpretability while preserving \ac{SCD} performance, even in low-dimensional settings.
In future work, we plan to extend the comparison to contextualised embeddings and multilingual \ac{SCD} tasks, in order to more broadly evaluate the trade-offs between interpretability and performance.

\section*{Limitations}
While our proposed method, \ac{SCD}Tour, demonstrates promising results, it has the following limitations.

First, our evaluations are conducted only for English, which is a morphologically limited language.
This is due to the necessity of both quantitative evaluation and qualitative analysis, which require extensive lexical resources and contextual understanding that are readily available for English. 
However, our proposed method is language-agnostic, and we expect that it would generalise to languages other than English.

Second, we focus exclusively on \acp{SWE} in this paper. 
As discussed in \autoref{subsec:exp_setting}, \acp{SWE} provide a single vector per word, enabling direct inspection of axis-level information, and has been adopted in prior work~\cite{sato-2022-wordtour,yamagiwa-etal-2024-axistour,musil-marecek-2024-exploring}. 
In contrast, \acp{CWE} encode multiple sense-aware vectors per token, making axis-level interpretation more challenging. 
While we prioritised interpretability in our analysis, \ac{CWE} often offers stronger performance in \ac{SCD}. 
A recent study has shown that \ac{SCD}-specific dimensions exist in \ac{CWE}~\cite{aida-bollegala-2025-investigating}, suggesting that future work could extend \ac{SCD}Tour to contextualised embeddings.

\section*{Ethical Considerations}
This paper does not introduce new datasets or models.
We conduct our experiments using existing datasets and pre-trained models.
To the best of our knowledge, no ethical issues have been reported regarding those evaluation datasets (SemEval-2020 Task 1 English~\cite{schlechtweg-etal-2020-semeval}, derived from CCOHA~\cite{alatrash-etal-2020-ccoha}).
Pre-trained models, such as GloVe and LLama-3.1, may contain social biases~\cite{kaneko-bollegala-2019-gender,basta-etal-2019-evaluating,oba-etal-2024-contextual}.
Future work should assess how these biases might be reflected in the obtained axes and affect interpretation in real-world applications.

\section*{Acknowledgements}
Taichi Aida would like to acknowledge the support by JST, the establishment of university fellowships towards the creation of science technology innovation, Grant Number JPMJFS2139.

\bibliography{anthology,custom}

\appendix
\section{Experimental Details}
\label{app_sec:details}

\subsection{Data Statistics}
\label{app_subsec:details_data}
To evaluate the effectiveness of the \ac{SCD}Tour, we performed a preliminary experiment using the Word Embedding Benchmarks, which include three subtasks: categorical, similarity, and analogy.
The benchmark is publicly available under the MIT License.\footnote{\url{https://github.com/kudkudak/word-embeddings-benchmarks}}

In the SemEval-2020 Task 1 for English, the dataset is constructed from the Cleaned Corpus of Historical American English (CCOHA)~\cite{alatrash-etal-2020-ccoha}, which contains time-separated newspapers, magazines, and (non-)fiction books suitable for analysing diachronic semantic change.\footnote{This dataset is licensed under the Creative Commons Attribution 4.0 International License.}
\autoref{tab:data} summarises the statistics of the lemmatised version of the corpora, including time periods, the number of target words and tokens.

\begin{table}[h]
    \centering
    \begin{tabular}{ccc} \toprule
        Time Period & \#Targets & \#Tokens \\ \midrule
        1810s--1860s & \multirow{2}{*}{37} & 6.5M \\
        1960s--2010s & & 6.7M \\ \bottomrule
    \end{tabular}
    \caption{Statistics of the \ac{SCD} benchmark, SemEval-2020 Task 1 (English). \#Targets and \#Tokens indicate the number of target words and tokens.}
    \label{tab:data}
\end{table}

\subsection{Hyperparameters}
\label{app_subsec:details_hyperparameters}
We used the pre-trained \texttt{GloVe 6B}\footnote{It is available at \url{https://nlp.stanford.edu/projects/glove/} under the Public Domain Dedication and License.} in the preliminary experiment.
For the main \ac{SCD} experiments, which aim to investigate how performance and interpretability can be maintained even after dimension reduction, we follow the procedure of \citet{laicher-etal-2021-explaining} for selecting hyperparameters for the \ac{SGNS} model.
Specifically, we perform a grid search over the values in \autoref{app_tab:hyperparameter}.\footnote{We used the LSCDetection toolkit available at \url{https://github.com/Garrafao/LSCDetection} . It is licensed under the GNU General Public License.}
For each configuration, we evaluate the model performance on the \ac{SCD} ranking task and select the best setting that yields the highest Spearman's correlation.
This hyperparameter tuning is performed jointly across different time periods.
\autoref{app_tab:hyperparameter} also shows the best setting, which is used in the main experiments presented in \autoref{subsec:exp_results}.

\begin{table}[h]
    \centering
    \begin{tabular}{cc} \toprule
        Parameter & Values \\ \midrule
        window size & 5, \textbf{10} \\
        dimension & 50, 100, \textbf{200}, 300 \\
        iteration & \textbf{5}, 10, 20, 30 \\
        negative samples & 5\\\bottomrule
    \end{tabular}
    \caption{Hyperparameters used for SGNS. \textbf{Bold} values indicates the best settings.}
    \label{app_tab:hyperparameter}
\end{table}

\subsection{Prompts}
\label{app_subsec:prompts}
\begin{figure}[t]
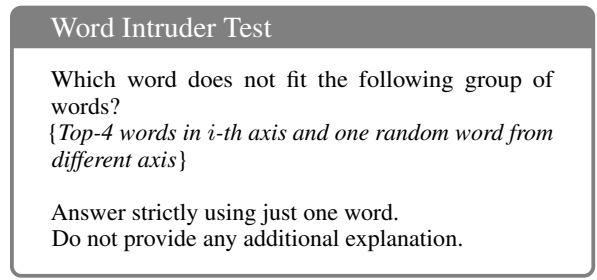

    \centering
    \begin{tcolorbox}[title=Word Intruder Test, fontupper=\small, colframe=gray, colback=white]
    Which word does not fit the following group of words? \\
    \{\textit{Top-4 words in $i$-th axis and one random word from different axis}\} \\
    
    Answer strictly using just one word. \\
    Do not provide any additional explanation.
    \end{tcolorbox}
    \caption{Prompt used for word intruder test. This prompt is referred to \citet{musil-marecek-2024-exploring}.}
    \label{fig:prompt_wit}
\end{figure}

\begin{figure}[t]
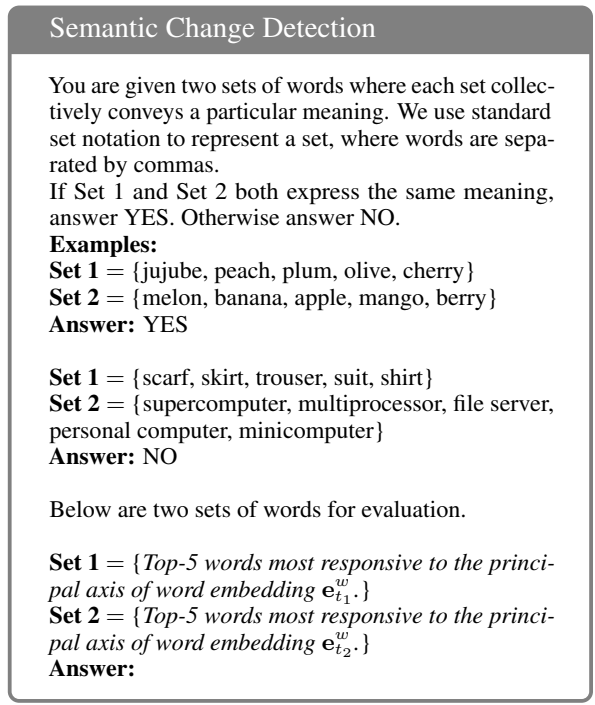

    \centering
    \begin{tcolorbox}[title=Semantic Change Detection, fontupper=\small, colframe=gray, colback=white]
    You are given two sets of words where each set collectively conveys a particular meaning.  
    We use standard set notation to represent a set, where words are separated by commas.

    If Set 1 and Set 2 both express the same meaning, answer YES. Otherwise answer NO.

    \textbf{Examples:}
    
    \textbf{Set 1} $=$ \{jujube, peach, plum, olive, cherry\} \\
    \textbf{Set 2} $=$ \{melon, banana, apple, mango, berry\} \\
    \textbf{Answer:} YES \\

    \textbf{Set 1} $=$ \{scarf, skirt, trouser, suit, shirt\} \\
    \textbf{Set 2} $=$ \{supercomputer, multiprocessor, file server, personal computer, minicomputer\} \\ 
    \textbf{Answer:} NO \\

    Below are two sets of words for evaluation. \\
    
    \textbf{Set 1} $=$ \{\textit{Top-5 words most responsive to the principal axis of word embedding $\mathbf{e}^{w}_{t_1}$.}\} \\
    \textbf{Set 2} $=$ \{\textit{Top-5 words most responsive to the principal axis of word embedding $\mathbf{e}^{w}_{t_2}$.}\} \\
    \textbf{Answer:} 
    \end{tcolorbox}
    \caption{Prompt used for semantic change detection (binary classification).}
    \label{fig:prompt_scd_binary}
\end{figure}

In the \ac{WIT} and \ac{SCD} binary task, we report the average accuracy over five runs to account for any variance in the generation process using three models: \texttt{Llama-3.1-8B-Instruct}, \texttt{Gemma-3-4b-it}, and \texttt{Qwen3-8B}\footnote{These models are licensed under the Llama 3.1 Community License, Gemma License, and Apache 2.0 License, respectively.}.
Prompts for each task are shown in \autoref{fig:prompt_wit} and \autoref{fig:prompt_scd_binary}.

\section{Additional \ac{SCD} Results}
\label{app_sec:scd}
In addition to the analysis presented in \autoref{sec:exp}, we provide further case studies on two word pairs: (i) \textit{plane} (shifted from \textit{(mathematical) surface} to \textit{aircraft}) and \textit{tree} (stable), and (ii) \textit{attack} (extended to include \textit{heart attack}) and \textit{relationship} (stable).
\autoref{tab:analysis_plane_tree} and \autoref{tab:analysis_attack_relationship} confirm that SCDTour can preserve the interpretability of semantic change while the dimension is reduced to $k = 20$.
We observe the following trends across all target words described in \autoref{sec:exp}.
At $k = 100$, all methods can retrieve representative words corresponding to the meaning of the target word.
SCDTour maintains the ability to extract the relevant words at $k = 20$, demonstrating robustness against baselines.
However, at $k = 5$, no method can correctly capture the meaning of the target word, due to excessive merging of distinct meanings into a single axis.

\begin{table*}[t]
    \centering
    \resizebox{\textwidth}{!}{
    \begin{tabular}{lrlrl} \toprule
        \multirow{2}{*}{Method} & \multicolumn{2}{c}{word: \textit{plane}} & \multicolumn{2}{c}{word: \textit{tree}}  \\ 
        & axis & words & axis & words \\ \midrule
        \multicolumn{5}{c}{$d = 200$ (full)} \\ \midrule
            \multirow{2}{*}{Raw} & \textcolor{black!50}{28} & \textcolor{black!50}{\textbf{z}, \textbf{q}, \textbf{g}, \textbf{k}, \textbf{h}} & \textcolor{black!50}{70} & \textcolor{black!50}{\textbf{flowering}, \textbf{pear}, \textbf{pendulous}, \textbf{deciduous}, \textbf{sycamore}} \\ 
             & 92 & \textbf{ship}, \textbf{aboard}, \textbf{crew}, \textbf{reconnaissance}, \textbf{passenger} & 70 & \textbf{shrub}, \textbf{beech}, \textbf{tree}, \textbf{leaf}, \textbf{dogwood} \\ \midrule
        \multicolumn{5}{c}{$k = 100$} \\ \midrule
            \multirow{2}{*}{PCA} & \textcolor{black!50}{99} & \textcolor{black!50}{1200, not, oakwood, lie, discovery} & \textcolor{black!50}{73} & \textcolor{black!50}{\textbf{tulip}, gouverneur, dignity, dreamy, sive} \\
             & 90 & perpetrate, honorably, forlom, somber, \textbf{voyager} & 73 & \$60, garibaldi, basilikon, winter, publication \\
            \multirow{2}{*}{ICA} & \textcolor{black!50}{70} & \textcolor{black!50}{pear, sycamore, shrubbery, tree, fern} & \textcolor{black!50}{70} & \textcolor{black!50}{\textbf{pear}, \textbf{sycamore}, \textbf{shrubbery}, \textbf{tree}, \textbf{fern}} \\
             & 92 & \textbf{ship}, \textbf{aboard}, uss, \textbf{passenger}, \textbf{reconnaissance} & 70 & \textbf{shrub}, \textbf{beech}, \textbf{cactus}, \textbf{tree}, \textbf{maple} \\
            \multicolumn{5}{l}{\textbf{SCDTour}} \\
            \multirow{2}{*}{\:$\lambda = 0.00$} & \textcolor{black!50}{43} & \textcolor{black!50}{\textbf{z}, \textbf{h}, \textbf{g}, \textbf{q}, \textbf{ky}} & \textcolor{black!50}{73} & \textcolor{black!50}{\textbf{tree}, \textbf{sycamore}, \textbf{pendulous}, \textbf{pear}, \textbf{deciduous}} \\
             & 56 & \textbf{ship}, \textbf{aboard}, \textbf{passenger}, \textbf{crew}, \textbf{air} & 73 & \textbf{shrub}, \textbf{leaf}, \textbf{beech}, \textbf{cactus}, \textbf{tree} \\ 
            \multirow{2}{*}{\:$\lambda = 0.25$} & \textcolor{black!50}{80} & \textcolor{black!50}{\textbf{z}, \textbf{abc}, \textbf{h}, \textbf{g}, \textbf{q}} & \textcolor{black!50}{41} & \textcolor{black!50}{\textbf{pear}, \textbf{elm}, hampshire, \textbf{shrub}, \textbf{flowering}} \\
             & 68 & \textbf{ship}, \textbf{crew}, \textbf{aboard}, \textbf{passenger}, uss & 41 & hampshire, broome, rochelle, powel, orleans \\
            \multirow{2}{*}{\:$\lambda = 1.00$} & \textcolor{black!50}{64} & \textcolor{black!50}{\textbf{z}, \textbf{abc}, \textbf{g}, \textbf{h}, \textbf{ky}} & \textcolor{black!50}{5} & \textcolor{black!50}{\textbf{flowering}, \textbf{sycamore}, \textbf{shrub}, \textbf{pear}, \textbf{herbaceous}} \\
            & 13 & \textbf{ship}, \textbf{aboard}, \textbf{flight}, uss, \textbf{crew} & 5 & \textbf{shrub}, \textbf{beech}, \textbf{cactus}, \textbf{deciduous}, \textbf{dogwood} \\ \midrule
        \multicolumn{5}{c}{$k = 20$} \\ \midrule
            \multirow{2}{*}{PCA} & \textcolor{black!50}{10} & \textcolor{black!50}{disbelief, god, almighty, forgiveness, whosoever} & \textcolor{black!50}{15} & \textcolor{black!50}{turner, protege, robertson, kidnapper, memphis} \\
             & 16 & sell, buy, bourse, outfit, fancy & 13 & defendant, squalid, humbert, detention, prison \\
            \multirow{2}{*}{ICA} & \textcolor{black!50}{10} & \textcolor{black!50}{utterly, pour, word, vegetation, russia} & \textcolor{black!50}{11} & \textcolor{black!50}{exempt, extricate, exemption, deviate, detract} \\
             & 14 & 7,000, gordon, dip, magareta, edge & 11 & derive, exempt, detract, emanate, refrain \\
            \multicolumn{5}{l}{\textbf{SCDTour}} \\
            \multirow{2}{*}{\:$\lambda = 0.00$} & \textcolor{black!50}{3} & \textcolor{black!50}{indictment, politician, lunatic, defendant, adjudge} & \textcolor{black!50}{14} & \textcolor{black!50}{guardianship, \textbf{twig}, \textbf{oak}, \textbf{bough}, \textbf{flowering}} \\
             & 11 & delighted, merge, sob, equip, inclined & 14 & owl, squirrel, assortment, \textbf{tree}, \textbf{conservation} \\ 
            \multirow{2}{*}{\:$\lambda = 0.25$} & \textcolor{black!50}{16} & \textcolor{black!50}{\textbf{z}, \textbf{q}, \textbf{w}, \textbf{g}, \textbf{arc}} & \textcolor{black!50}{8} & \textcolor{black!50}{\textbf{pear}, \textbf{vine}, \textbf{tree}, \textbf{elm}, \textbf{pendulous}} \\
             & 13 & earnings, liquidate, \textbf{regulator}, depositor, underwrite & 8 & broome, hampshire, mistake, soon, \textbf{fern} \\
            \multirow{2}{*}{\:$\lambda = 1.00$} & \textcolor{black!50}{12} & \textcolor{black!50}{\textbf{z}, \textbf{g}, \textbf{ky}, \textbf{q}, \textbf{sm}} & \textcolor{black!50}{1} & \textcolor{black!50}{\textbf{hemlock}, stunted, moneywort, \textbf{crop}, \textbf{apple-tree}} \\
            & 9 & \textbf{drawbridge}, progress, \textbf{aerial}, lu, episode & 1 & earnings, depositor, \textbf{shrub}, dividend, seller \\ \midrule
        \multicolumn{5}{c}{$k = 5$} \\ \midrule
            \multirow{2}{*}{PCA} & \textcolor{black!50}{4} & \textcolor{black!50}{pie, mince, first-rate, stuff, tight} & \textcolor{black!50}{3} & \textcolor{black!50}{glasgow, 1835, 43, sloop, 1830} \\
             & 3 & kentucky, oakland, 153, md, fl & 3 & kentucky, oakland, 153, md, fl \\
            \multirow{2}{*}{ICA} & \textcolor{black!50}{3} & \textcolor{black!50}{qui, je, vous, come, zo} & \textcolor{black!50}{2} & \textcolor{black!50}{vanity, emotion, apprehension, temporary, petty} \\
             & 2 & joy, anger, disappointment, anxiety, sorrow & 3 & sus, queue, por, la, como \\
            \multicolumn{5}{l}{\textbf{SCDTour}} \\
            \multirow{2}{*}{\:$\lambda = 0.00$} & \textcolor{black!50}{2} & \textcolor{black!50}{spec, r, ben, heave, brimstone} & \textcolor{black!50}{3} & \textcolor{black!50}{militia, thornton, assistance, impression, opportunity} \\
             & 2 & sussex, premature, deck, 262, grunt & 3 & enormous, specimen, revelation, outstretched, handkerchief \\ 
            \multirow{2}{*}{\:$\lambda = 0.25$} & \textcolor{black!50}{0} & \textcolor{black!50}{consistent, lunatic, facility, pikes, unacquainted} & \textcolor{black!50}{2} & \textcolor{black!50}{laboratory, deduce, imaginary, varmint, jury} \\
             & 3 & lecture, oatmeal, dark, objective, industry & 2 & clearly, greased, \textbf{cultivated}, legitimate, hamlet \\ 
            \multirow{2}{*}{\:$\lambda = 1.00$} & \textcolor{black!50}{3} & \textcolor{black!50}{cambridge, xx\_v, aimed, mocha, tut} & \textcolor{black!50}{0} & \textcolor{black!50}{web, whereby, resounding, bind, last} \\
            & 3 & mystery, biblical, z, dairy, fais & 0 & unison, credible, anticipated, \$800, dramatically \\ \bottomrule
    \end{tabular}
    }
    \caption{Representative words from the most activated axis of the target word embedding (\textit{plane} and \textit{tree}) at two time periods $t_1$ (shown in \textcolor{black!50}{gray}) and $t_2$ (in black), across different methods. For each method and target word, we identify the axis with the highest value in the embedding, and list the top-$5$ words associated with that axis. Words that reflect the meaning of the target word are highlighted in \textbf{bold}. \textit{Plane} underwent a semantic change (\textit{(mathematical) surface} to \textit{aircraft}), while \textit{tree} remained stable. At $d = 200$ and $k = 100$, all methods capture the corresponding meanings for both time periods. \ac{SCD}Tour ($\lambda = 0.25$) maintains this interpretability even at $k = 20$, whereas other methods such as PCA or ICA struggle. At $k = 5$, no method successfully preserves corresponding meanings due to the limited number of axes or excessive merging.}
    \label{tab:analysis_plane_tree}
\end{table*}

\begin{table*}[t]
    \centering
    \resizebox{\textwidth}{!}{
    \begin{tabular}{lrlrl} \toprule
        \multirow{2}{*}{Method} & \multicolumn{2}{c}{word: \textit{attack}} & \multicolumn{2}{c}{word: \textit{relationship}}  \\ 
        & axis & words & axis & words \\ \midrule
        \multicolumn{5}{c}{$d = 200$ (full)} \\ \midrule
            \multirow{2}{*}{Raw} & \textcolor{black!50}{49} & \textcolor{black!50}{amends, \textbf{arrangement}, \textbf{debut}, \textbf{effort}, \textbf{appearance}} & \textcolor{black!50}{34} & \textcolor{black!50}{\textbf{aryan}, \textbf{discrepancy}, clavicle, \textbf{demarcation}, \textbf{estrangement}} \\ 
             & 157 & \textbf{respiratory}, \textbf{intestinal}, \textbf{pulmonary}, \textbf{uterine}, \textbf{inflammation} & 34 & \textbf{correlation}, \textbf{disparity}, \textbf{relationship}, \textbf{discrepancy}, \textbf{distinction} \\ \midrule
            \multicolumn{5}{c}{$k = 100$} \\ \midrule
            \multirow{2}{*}{PCA} & \textcolor{black!50}{93} & \textcolor{black!50}{\textbf{alert}, \textbf{fated}, seventh, ihe, \textbf{resentment}} & \textcolor{black!50}{94} & \textcolor{black!50}{globe, self-satisfied, isaiah, dainty, \textbf{area}} \\
             & 77 & declaration, \textbf{throbbing}, holland, i'he, weatherworn & 69 & \textbf{friendship}, bridge, \textbf{bond}, ingly, self-denying \\
            \multirow{2}{*}{ICA} & \textcolor{black!50}{70} & \textcolor{black!50}{pear, sycamore, shrubbery, tree, fern} & \textcolor{black!50}{51} & \textcolor{black!50}{\textbf{fellow-man}, protege, pursuer, townsman, fellow-citizen} \\
             & 54 & minister, prime, beautiful, mountain, marriage & 71 & accordance, \textbf{sympathize}, \textbf{interfere}, \textbf{coincide}, \textbf{comply} \\
            \multicolumn{5}{l}{\textbf{SCDTour}} \\
            \multirow{2}{*}{\:$\lambda = 0.00$} & \textcolor{black!50}{90} & \textcolor{black!50}{rely, dependent, \textbf{encroach}, devolve, \textbf{preye}} & \textcolor{black!50}{19} & \textcolor{black!50}{\textbf{aryan}, \textbf{russia}, \textbf{discrepancy}, \textbf{austria}, \textbf{sweden}} \\
             & 90 & embark, \textbf{verge}, rely, depending, reliance & 19 & \textbf{austria}, \textbf{scandinavia}, \textbf{belgium}, \textbf{albania}, \textbf{ethiopia} \\ 
            \multirow{2}{*}{\:$\lambda = 0.25$} & \textcolor{black!50}{42} & \textcolor{black!50}{\textbf{arrangement}, amends, \textbf{appearance}, confession, \textbf{debut}} & \textcolor{black!50}{57} & \textcolor{black!50}{\textbf{aryan}, \textbf{discrepancy}, \textbf{russia}, clavicle, \textbf{scandinavian}} \\
             & 92 & \textbf{pulmonary}, \textbf{respiratory}, \textbf{infection}, \textbf{colon}, \textbf{liver} & 57 & \textbf{austria}, \textbf{scandinavia}, \textbf{belgium}, \textbf{albania}, 1715 \\
            \multirow{2}{*}{\:$\lambda = 1.00$} & \textcolor{black!50}{52} & \textcolor{black!50}{\textbf{artillery}, \textbf{cavalry}, \textbf{regiment}, \textbf{troop}, \textbf{infantry}} & \textcolor{black!50}{2} & \textcolor{black!50}{shrill, peal, reverberate, tinkling, dirge} \\
            & 22 & embark, dote, depending, depend, reliance & 2 & shrill, \textbf{correlation}, cymbal, muffle, hoarse \\ \midrule
        \multicolumn{5}{c}{$k = 20$} \\ \midrule
            \multirow{2}{*}{PCA} & \textcolor{black!50}{12} & \textcolor{black!50}{\textbf{camp}, appetite, swallow, draught, pipe} & \textcolor{black!50}{19} & \textcolor{black!50}{\textbf{banter}, data, \textbf{affinity}, \textbf{disinclination}, meaning} \\
             & 11 & polka, profane, trickster, wilde, gobbler & 18 & underlie, \textbf{collapse}, sweaty, na, ta \\
            \multirow{2}{*}{ICA} & \textcolor{black!50}{12} & \textcolor{black!50}{god, cook, thus, throw, convincing} & \textcolor{black!50}{10} & \textcolor{black!50}{utterly, pour, word, vegetation, \textbf{russia}} \\
             & 10 & chain, atmosphere, interview, prove, phrase & 10 & chain, \textbf{atmosphere}, interview, prove, phrase \\
            \multicolumn{5}{l}{\textbf{SCDTour}} \\
            \multirow{2}{*}{\:$\lambda = 0.00$} & \textcolor{black!50}{14} & \textcolor{black!50}{\textbf{guardianship}, twig, oak, bough, flowering} & \textcolor{black!50}{10} & \textcolor{black!50}{neither, nor, plentitude, \textbf{neighbourhood}, \textbf{homelike}} \\
             & 2 & \textbf{transplant}, \textbf{inflammation}, \textbf{infection}, \textbf{disorder}, \textbf{respiratory} & 3 & winner, defendant, moslem, pennant, congressmen \\ 
            \multirow{2}{*}{\:$\lambda = 0.25$} & \textcolor{black!50}{14} & \textcolor{black!50}{d'etre, de, cet, \textbf{normandie}, rien} & \textcolor{black!50}{10} & \textcolor{black!50}{\textbf{abstain}, deduce, \textbf{exempt}, lately, alleghany} \\
             & 14 & pasado, muy, ramn, quelque, misma & 11 & \textbf{tension}, \textbf{belgium}, \textbf{confederation}, \textbf{unrest}, \textbf{albania} \\
            \multirow{2}{*}{\:$\lambda = 1.00$} & \textcolor{black!50}{4} & \textcolor{black!50}{ruminate, extremely, lavish, depending, preye} & \textcolor{black!50}{3} & \textcolor{black!50}{meantime, \textbf{assist}, meanwhile, inmost, reciprocate} \\
            & 19 & jamieson, jamie, galbraith, oliver, shrewsbury & 0 & 2,200, audible, sidle, syllable, difference \\ \midrule
        \multicolumn{5}{c}{$k = 5$} \\ \midrule
            \multirow{2}{*}{PCA} & \textcolor{black!50}{3} & \textcolor{black!50}{\textbf{glasgow}, 1835,  43, \textbf{sloop}, 1830} & \textcolor{black!50}{2} & \textcolor{black!50}{mortimer, digby, pauline, harding, terence} \\
             & 3 & Kentucky, oakland, 153, md, fl & 4 & tasty, pill, prescription, medication, dessert \\
            \multirow{2}{*}{ICA} & \textcolor{black!50}{3} & \textcolor{black!50}{qui, je, vous, come, zo} & \textcolor{black!50}{2} & \textcolor{black!50}{vanity, emotion, apprehension, temporary, petty} \\
             & 2 & joy, anger, disappointment, \textbf{anxiety}, sorrow & 2 & joy, anger, disappointment, anxiety, sorrow \\
            \multicolumn{5}{l}{\textbf{SCDTour}} \\
            \multirow{2}{*}{\:$\lambda = 0.00$} & \textcolor{black!50}{3} & \textcolor{black!50}{\textbf{militia}, thornton, \textbf{assistance}, impression, \textbf{opportunity}} & \textcolor{black!50}{2} & \textcolor{black!50}{spec, r, ben, heave, brimstone} \\
             & 0 & ciudad, mrs, cell, marietta, breed & 0 & ciudad, mrs, cell, marietta, breed \\ 
            \multirow{2}{*}{\:$\lambda = 0.25$} & \textcolor{black!50}{2} & \textcolor{black!50}{\textbf{laboratory}, deduce, imaginary, varmint, jury} & \textcolor{black!50}{2} & \textcolor{black!50}{laboratory, deduce, imaginary, \textbf{varmint}, jury} \\
             & 0 & gobble, chilton, consecutive, convenience, twenty-five & 1 & 98, nonsense, perspective, havin, balk \\
            \multirow{2}{*}{\:$\lambda = 1.00$} & \textcolor{black!50}{1} & \textcolor{black!50}{conjugal, brightest, saturday, consider, greenish} & \textcolor{black!50}{0} & \textcolor{black!50}{web, whereby, resounding, bind, last} \\
            & 4 & gibson, brighten, colorless, coarse, ol & 0 & unison, credible, anticipated, \$800, dramatically \\ \bottomrule
    \end{tabular}
    }
    \caption{Representative words from the most activated axis of the target word embedding (\textit{attack} and \textit{relationship}) at two time periods $t_1$ (shown in \textcolor{black!50}{gray}) and $t_2$ (in black), across different methods. For each method and target word, we identify the axis with the highest value in the embedding, and list the top-$5$ words associated with that axis. Words that reflect the meaning of the target word are highlighted in \textbf{bold}. \textit{Attack} exhibits a semantic shift through the inclusion of the medical sense (\textit{heart attack}) in the later time period $t_2$, whereas \textit{relationship} remains semantically stable. At $d = 200$ and $k = 100$, most methods correctly capture relevant meanings for both time periods. At $k = 20$, only \ac{SCD}Tour ($\lambda = 0.00, 0.25$) consistently identifies axes aligned with the new sense of \textit{attack}, while other methods retrieve more unrelated terms. At $k = 5$, most methods fail to reflect either meaning due to the limited number of axes or excessive axis merging.}
    \label{tab:analysis_attack_relationship}
\end{table*}

\section{Prompt Design and LLM Behaviour}
\label{app_sec:prompt}
While zero-shot prompts worked well for \ac{WIT}, they did not yield meaningful outputs in the \ac{SCD} binary classification task: the \ac{LLM} consistently returned YES or NO regardless of the input sets.
To address this issue, and following \citet{sorensen-etal-2022-information}, we designed few-shot prompts using WordNet\footnote{\url{https://www.nltk.org/howto/wordnet.html}} synsets.
We constructed both strong positive and negative examples from these synsets, as shown in \autoref{fig:prompt_scd_binary}.
We found that these few-shot prompts help the \ac{LLM} toward recognising subtle semantic differences between given sets of words, mitigating their strict similarity thresholds observed in \citet{sorensen-etal-2022-information}.

In addition to the prompting strategy, we observed that both instruction tuning and explicit instructions were essential for the \ac{WIT} and \ac{SCD}.
Without instruction-tuned models such as \texttt{meta-llama/Llama-3.1-8B}, the \ac{LLM} often failed to follow the task setup.
Even with instruction-tuned models, ambiguous prompts lead \acp{LLM} to return generic outputs (e.g. ``Sure!'').
To mitigate this issue, we included clear phrases (e.g. ``Answer strictly using just one word.''), which improved consistency.

\end{document}